  \documentclass{gretsi}
 \usepackage[latin1]{inputenc}   
\usepackage[english,french]{babel}   
 \usepackage{times}			
 
\usepackage{amssymb}
\usepackage{bm}
\usepackage{subfigure}
\usepackage{amsmath,graphicx,dsfont,color}
\usepackage{amsfonts}
\usepackage{amssymb}
\usepackage{array}
\usepackage{algorithm}
\usepackage{algorithmic}
\usepackage{placeins}
\usepackage{lipsum}
 
\newcommand{\squeezeup}{\vspace{-2.5mm}}
\newcommand{\squeezeupsmall}{\vspace{-1mm}}

\newcommand{\nsize}{n}
\newcommand{\node}{v}
\newcommand{\realset}{\mathbb{R}}
\newcommand{\I}{\mathbf{I}}
\newcommand{\graph}{\mathcal{G}}
\newcommand{\mst}{\mathcal{APM}}
\newcommand{\msf}{\mathcal{F}}
\newcommand{\nodes}{\mathcal{V}}
\newcommand{\edges}{\mathcal{E}}
\newcommand{\partition}{\mathsf{\pi}}
\newcommand{\region}{\mathsf{R}}
\newcommand{\hierarchy}{\mathcal{H}}

\newcommand{\fultra}{\bm{\thUCM}}
\newcommand{\W}{\mathbf{W}}

\newcommand{\probamap}{\mathrm{CP}}
\newcommand{\thUCM}{\lambda}

\newcommand{\tree}{$T$} 

\titre{Segmentation hiÈrarchique faiblement supervisÈe}

\auteur{\coord{Amin}{Fehri}{},
        \coord{Santiago}{Velasco-Forero}{},
    \coord{Fernand}{Meyer}{}}

\adresse{\affil{}{Mines ParisTech, PSL Research University, Centre de Morphologie MathÈmatique,  \\
         35 rue Saint-HonorÈ, 77305 Fontainebleau Cedex, France}}

\email{\{amin.fehri,
santiago.velasco,
fernand.meyer@mines-paristech.fr\}}

\resumefrancais{La segmentation d'images est le processus visant ‡ partitionner une image en rÈgions d'intÈrÍt par rapport ‡ des critËres donnÈs. La segmentation hiÈrarchique s'est rÈvÈlÈe Ítre une approche majeure en ce sens, car elle favorise l'Èmergence des rÈgions importantes ‡ diffÈrentes Èchelles. D'autre part, de nombreuses mÈthodes nous permettent aujourd'hui d'avoir une information a priori sur la position des structures d'intÈrÍt dans les images. Dans cet article, nous proposons un algorithme versatile de segmentation hiÈrarchique qui permet de prendre en compte une telle information a priori pour donner une segmentation hiÈrarchique qui met l'accent sur les rÈgions ou contours d'intÈrÍt tout en prÈservant les structures importantes dans l'image. Une application au problËme de segmentation faiblement supervisÈe est prÈsentÈe.}

\resumeanglais{
Image segmentation is the process of partitioning an image into a set of meaningful regions according to some criteria. Hierarchical segmentation has emerged as a major trend in this regard as it favors the
emergence of important regions at different scales. On the other hand, many methods allow us to have prior information on the position of structures of interest in the images. In this paper, we present a versatile hierarchical segmentation method that takes into account any prior spatial information and outputs a hierarchical segmentation that emphasizes the contours or regions of interest while preserving the important structures in the image. An application of this method to the weakly-supervised segmentation problem is presented.}

\begin{document}
\maketitle

\section{Introduction}

\squeezeup
\squeezeup

Dans cet article, nous proposons une mÈthode permettant de tirer parti de toute information spatiale prÈalablement obtenue sur une image pour obtenir une segmentation hiÈrarchique de cette image qui met en exergue ses rÈgions d'intÈrÍt. Ceci nous permet d'avoir plus de dÈtails dans ces rÈgions tout en conservant les informations structurelles fortes de l'image. 

\begin{figure}
\begin{center}
\includegraphics[width=0.85 \columnwidth]{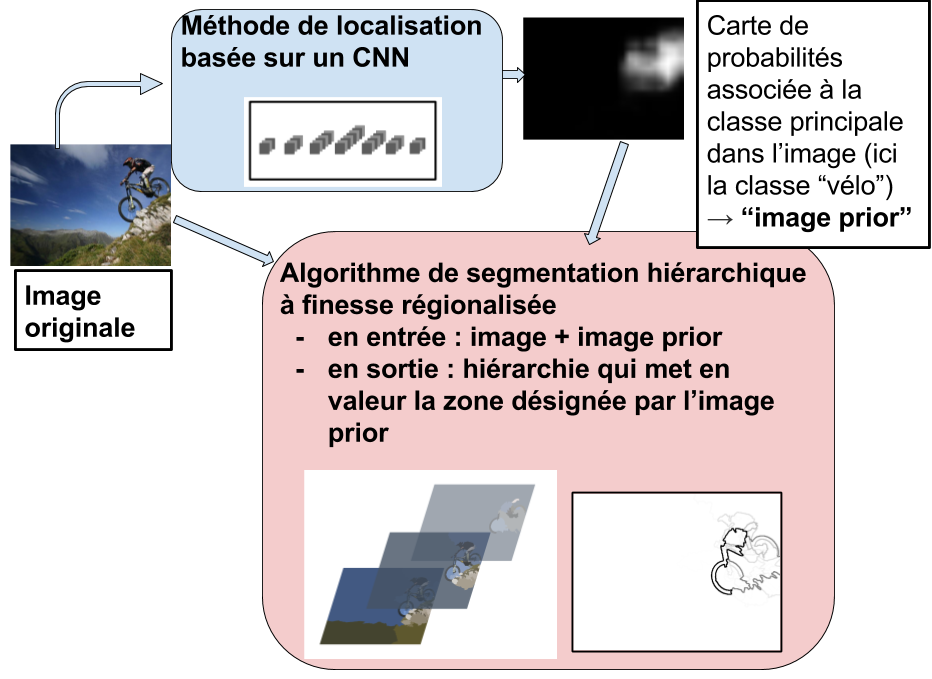}
\caption{Vue d'ensemble de la mÈthode proposÈe.}\label{Fig:Overview}
\end{center}
\squeezeup
\squeezeup
\squeezeup
\end{figure}

Les potentielles applications sont nombreuses. Cela permettrait par exemple, lorsqu'on a une capacitÈ de stockage limitÈe (e.g. pour de trËs larges images), de conserver les dÈtails de l'image dans les rÈgions d'intÈrÍt en prioritÈ. De nombreux cas Ètant abordÈs dans un article associÈ \cite{fehri17}, nous illustrerons dans la suite cette mÈthode uniquement par son application ‡ un problËme de segmentation faiblement supervisÈ. Une vue d'ensemble est proposÈe en figure \ref{Fig:Overview}.

Il a ÈtÈ montrÈ que la segmentation est de faÁon inhÈrente un problËme multi-Èchelles. 
C'est pourquoi la segmentation hiÈrarchique est devenue une tendance majeure ces derniËres annÈes, et qu'elle a donnÈ la plupart des algorithmes de l'Ètat de l'art actuel en segmentation. L'idÈe est de pas renvoyer uniquement une partition des pixels de l'image mais une structure multi-Èchelles qui rende compte des objets d'intÈrÍt ‡ toutes les Èchelles. Nous prÈsentons ici un algorithme de segmentation hiÈrarchique qui se concentre sur certaines zones prÈdÈterminÈes de l'image. L'aspect hiÈrarchique permet alors de choisir de faÁon simple le niveau de dÈtails dÈsirÈ en fonction de l'application.
De plus, cet algorithme est trËs versatile car l'information spatiale a priori qu'il prend en entrÈe peut Ítre obtenu par n'importe laquelle des nombreuses approches de vision par ordinateur existantes pour localiser les objets. De ce point de vue, ce travail s'inscrit dans le cadre des recherches visant ‡ incorporer un savoir prÈ-existant sur l'image pour la segmentation.

Dans la section \ref{sec:Hierarchies}, nous expliquerons comment construire et utiliser un algorithme de segmentation hiÈrarchique sur graphes. Puis nous prÈciserons dans la section \ref{sec:HierarchiesWithPrior} comment utiliser une information spatiale a priori sur un objet d'intÈrÍt pour obtenir une hiÈrarchie ‡ finesse rÈgionalisÈe. Enfin, un exemple d'application ‡ la segmentation faiblement supervisÈe sera prÈsentÈ dans la partie \ref{sec:CNNprior} et des conclusions seront tirÈes dans la partie \ref{sec:Conclusion}.


\squeezeup
\squeezeup

\section{HiÈrarchies et partitions}
\label{sec:Hierarchies}
\squeezeup
\subsection{Segmentation hiÈrarchique fondÈe sur des graphes}

Obtenir de faÁon directe une segmentation adaptÈe ‡ partir d'une image est trËs difficile. C'est pourquoi on fait souvent appel ‡ des hiÈrarchies pour organiser et proposer des contours intÈressants en les valuant. Dans cette section, nous rappelons au lecteur comment construire et utiliser des outils de segmentation hiÈrarchique sur graphes.

On suppose que, pour chaque image, une partition fine est obtenue par une segmentation initiale et qu'elle contient tous les contours qui font sens dans l'image. On dÈfinit alors une mesure de dissimilaritÈ entre tuiles adjacentes de cette partition fine. On peut alors voir l'image comme un graphe, le \textit{graphe d'adjacence de rÈgions} (GAR), dans lequel chaque noeud reprÈ\-sente une tuile de la partition fine; une arÍte relie deux noeuds si les rÈgions correspondantes sont voisines dans la partition fine; le poids de chaque arÍte est Ègal ‡ la dissimilaritÈ entre ces rÈgions. Travailler sur le GAR est beaucoup plus efficient que de travailler sur l'image, car il y a beaucoup moins de noeuds dans le GAR qu'il n'y a de pixels dans l'image.

De faÁon formelle, on note ce graphe $\graph=(\nodes,\edges,\W)$, o˘ $\nodes$ correspond au domaine de l'image ou ‡ l'ensemble des pixels ou rÈgions fines, $\edges \subset \nodes \times \nodes$ est l'ensemble des arÍtes reliant les rÈgions voisines, $\W: \edges \to \realset^{+}$ est la valeur de dissimilaritÈ gÈnÈralement basÈe sur une mesure du gradient local (ou couleur ou texture), par exemple $\W(i,j) \propto |\I(\node_i)-\I(\node_j)|$ avec $\I:\nodes \to \realset$ reprÈsentant l'intensitÈ de l'image.

L'arÍte reliant les noeuds $p$ et $q$ est notÈe $e_{pq}$. Un \textit{chemin} est une suite de noeuds et d'arÍtes : par exemple, un chemin liant les noeuds $p$ et $q$ est l'ensemble $\{p, e_{pt} , t, e_{ts}, s\}$. Un \textit{graphe connectÈ} est un sous-graphe dans lequel toute paire de noeuds est connectÈe par un chemin. Un \textit{cycle} est un chemin dont les extremitÈs coÔncident. Un \textit{arbre} est un graphe connectÈ sans cycle. Un \textit{arbre couvrant} est un arbre contenant tous les noeuds. Un \textit{arbre de poids minimum} (APM) $\mst(\graph)$ d'un graphe $\graph$ est un arbre couvrant avec le poids minimal possible, obtenu par exemple en utilisant l'algorithme de Boruvka (le poids d'un arbre Ètant Ègal ‡ la somme des poids de ses arÍtes). Une \textit{forÍt} est un ensemble d'arbres.

Une \emph{partition} $\partition$ d'un ensemble $\nodes$ est une collection de sous-ensembles de $\nodes$, tels que l'ensemble $\nodes$ soit l'union disjointe des sous-ensembles dans la partition, i.e. $\partition=\{\region_1,\region_2,\ldots,\region_k\}$ tel que : $\forall i, \region_i \subseteq \nodes \ $; $ \forall i\neq j,  \region_i \cap\region_j = \emptyset$ ; $\bigcup_{i}^{k}\region_i = \nodes$.

Couper toutes les arÍtes de $\mst(\graph)$ ayant une valuation supÈrieure ‡ un seuil $\thUCM$ mËne ‡ une forÍt de poids minimum (FPM) $\msf(\graph)$, i.e. ‡ une partition du graphe. La partition obtenue est alors la mÍme que celle qui aurait ÈtÈ obtenue en coupant les arÍtes de valuations supÈrieures ‡ $\thUCM$ directement dans $\graph$. 
Puisque travailler sur un $\mst(\graph)$ est moins co˚teux et fournit des rÈsultats similaires pour la segmentation, nous travaillons dans la suite uniquement avec un $\mst(\graph)$.


Couper les arÍtes par valuation dÈcroissante mËne donc ‡ une \emph{hiÈrarchie de partitions indexÈe} $(\hierarchy,\fultra)$, avec $\hierarchy$ une \emph{hiÈrarchie de partitions} i.e. une chaine de partitions imbriquÈes $\hierarchy=\{\partition_0, \partition_1,\ldots, \partition_\nsize| \forall j,k, \quad 0 \quad \leq j\leq k\leq \nsize \Rightarrow \partition_j \sqsubseteq \partition_k\}$, avec $\partition_\nsize$ la partition ‡ une seule rÈgion et $\partition_0$ la partition fine de l'image, et $\fultra: \hierarchy \to \realset^+$ Ètant un indice de stratification vÈrifiant $\fultra(\partition) < \fultra(\partition')$ pour deux partitions imbriquÈes $\partition \subset \partition'$. Cette application croissante nous permet d'attribuer ‡ chaque tronÁon de contour le niveau de la hiÈrarchie pour lequel il disparaÓt : c'est la \emph{saillance} du contour et on considËre que plus la saillance est grande, plus le contour est fort.
Pour une hiÈrarchie donnÈe, l'image dans laquelle chaque contour prend comme valeur sa saillance est appelÈe \textit{Carte de Contours UltramÈtriques} (CCU)\cite{arbelaez2011contour}. ReprÈsenter une hiÈrarchie par sa CCU est une faÁon simple d'avoir une idÈe de son effet car seuiller une CCU aboutit toujours ‡ un ensemble de courbes fermÈes et donc ‡ une partition. Dans cet article, pour une mei\-lleure visibilitÈ, nous reprÈsentons les CCU avec un contraste inversÈ.

Il y a plusieurs faÁons d'obtenir une segmentation ‡ partir d'une hiÈrarchie : (i) de faÁon simple en seuillant les saillances les plus hautes; (ii) en marquant certains noeuds comme importants et en calculant alors une partition conformÈment ‡ ces marqueurs, connu sous le nom de \textit{segmentation avec marqueurs}; (iii) en recherchant la segmentation minimisant une fonctionnelle ÈnergÈtique dÈfinie sur le graphe.

Or, la qualitÈ des segmentations obtenues dÈpend grandement de la dissimilaritÈ que nous utilisons, et en changer peut mener ‡ des segmentations plus pertinentes. La ligne de partage des eaux stochastique est un des outils possibles nous permettant d'attribuer aux contours des poids prenant en compte une information plus riche qu'une simple dissimilaritÈ locale, de telle sorte que des mÈthodes plus simples soient suffisantes pour en extraire des segmentations intÈressantes.


\squeezeup
\squeezeup
\subsection{HiÈrarchies de Ligne de Partage des Eaux Stochastique (LPES)}

La ligne de partage des eaux stochastique (LPES), introduite dans \cite{angulo2007stochastic} par simulation et Ètendue ‡ une approche par graphes dans \cite{meyer15}, est un outil versatile pour construire des hiÈrarchies. L'idÈe originale est d'opÈrer ‡ de multiples reprises une segmentation avec marqueurs avec des marqueurs alÈatoires puis de valuer chaque arÍte de l'$\mst$ par sa frÈquence d'apparition dans les segmentations rÈsultantes. 

En effet, en distribuant des marqueurs sur le GAR  $\graph$, on peut construire une segmentation comme une FPM $\msf(\graph)$ dans laquelle chaque arbre prend racine dans un noeud marquÈ. On constate que la segmentation avec marqueurs est possible directement sur l'$\mst$ : on doit alors couper, pour chaque paire de marqueurs, l'arÍte la plus haute sur le chemin les reliant. De plus, il y a un domaine de variation dans lequel chaque marqueur peut se dÈplacer tout en aboutissant ‡ la mÍme segmentation. Plus de dÈtails sont fournis dans l'article associÈ \cite{fehri17}.


ConsidÈrons sur l'$\mst$ une arÍte $e_{st}$ de poids $\omega_{st}$ et calculons sa probabilitÈ d'Ítre coupÈe. Nous coupons toutes les arÍtes de l'$\mst$ ayant un poids supÈrieur ou Ègal ‡ $\omega_{st}$, ce qui aboutit ‡ deux arbres $\tree_{s}$ et $\tree_{t}$ de racines $s$ et $t$. Si au moins un marqueur tombe dans le domaine $\region_{s}$ des noeuds de $\tree_{s}$ et au moins un marqueur tombe dans le domaine $\region_{t}$ des noeuds de $\tree_{t}$, alors $e_{st}$ sera coupÈ dans la segmentation finale.

Si l'on note $\mu(\region)$ le nombre de marqueurs alÈatoires tombant dans la rÈgion $\region$, nous voulons attribuer ‡ $e_{st}$ la valeur de probabilitÈ suivante :
\squeezeupsmall
\begin{equation} \label{Proba}
\begin{split}
\tilde{\omega}_{st} & = \mathbb{P}[(\mu(\region_{s}) \geq 1) \land (\mu(\region_{t}) \geq 1)] \\
& = 1-\mathbb{P}[(\mu(\region_{s}) = 0) \lor (\mu(\region_{t}) = 0)] \\
& = 1-\mathbb{P}(\mu(\region_{s}) = 0) -\mathbb{P}(\mu(\region_{t}) = 0) \\
& +\mathbb{P}(\mu(\region_{s} \cup \region_{t}) = 0) \\
\end{split}
\end{equation}
\squeezeupsmall
Si les marqueurs sont distribuÈs suivant une distribution de Poisson, alors pour une rÈgion $R$:
\begin{equation} \label{Poisson}
\mathbb{P}(\mu(R) = 0)=\exp^{-\Lambda(R)}, 
\end{equation}
avec $\Lambda(R)$ l'espÈrance du nombre de marqueurs tombant dans $R$. La probabilitÈ devient donc :
\begin{equation} \label{newProba}
\begin{split}
\tilde{\omega}_{st} = 1-\exp^{-\Lambda(\region_{s})}-\exp^{-\Lambda(\region_{t})} 
+\exp^{-\Lambda(\region_{s} \cup \region_{t})}
\end{split}
\end{equation}

Avec, lorsque la distribution de Poisson a une densitÈ uniforme $\lambda$:
\squeezeup
\begin{equation} \label{uniform}
\Lambda(R) = \texttt{aire}(R) \lambda, 
\end{equation}
Et lorsque la distribution de Poisson a une densitÈ non-uniforme $\lambda$:
\squeezeup
\begin{equation} \label{nonuniform}
\Lambda(R)=\int_{(x,y) \in R} \lambda(x,y) \, \mathrm dx \mathrm dy
\end{equation}

La sortie de l'algorithme de LPES dÈpend donc de l'$\mst$ de dÈpart (structure et valuations des arÍtes) et de la loi de probabilitÈ gouvernant la distribution des marqueurs. De plus, les hiÈrarchies de LPES peuvent Ítre chaÓnÈes puisque les nouvelles valuations des contours correspondent ‡ des nouvelles valeurs pour l'$\mst$ et que l'on peut alors rÈitÈrer le processus dÈcrit prÈcedemment sur cet $\mst$. Ceci amËne un large espace d'exploration qui peut Ítre utilisÈ dans une chaÓne de segmentation \cite{fehri16}.

Du fait de sa versatilitÈ et de ses bonnes performances, la LPES reprÈsente un bon algorithme de dÈpart ‡ modifier pour y injecter une information spatiale a priori. En effet, lorsque l'on a une telle information sur l'image, est-il possible de l'utiliser pour avoir plus de dÈtails dans certaines parties de l'image plutÙt que dans d'autres ? Cela permettrait de simplifier et d'amÈ\-liorer des traitements ultÈrieurs en concentrant les efforts dans les zones d'intÈrÍt.

\squeezeup
\squeezeup
\section{HiÈrarchies avec information a priori}
\label{sec:HierarchiesWithPrior}
\squeezeup
\subsection{HiÈrarchie ‡ finesse rÈgionalisÈe (HFR)}
\label{ssec:HFR}

Dans la LPES originale, une distribution uniforme de marqueurs est utilisÈe (quelque soit leur taille ou leur forme). Pour avoir des contours plus forts dans une rÈgion spÈcifique de l'image, on propose d'adapter le modËle pour que plus de marqueurs soient distribuÈs dans cette rÈgion.

\begin{figure}
\begin{center}
\subfigure[]{\includegraphics[width=.225\columnwidth]{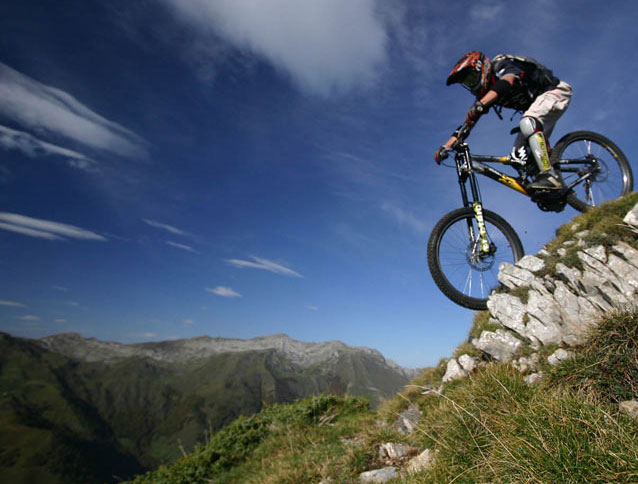}}
\subfigure[]{\includegraphics[width=.225\columnwidth]{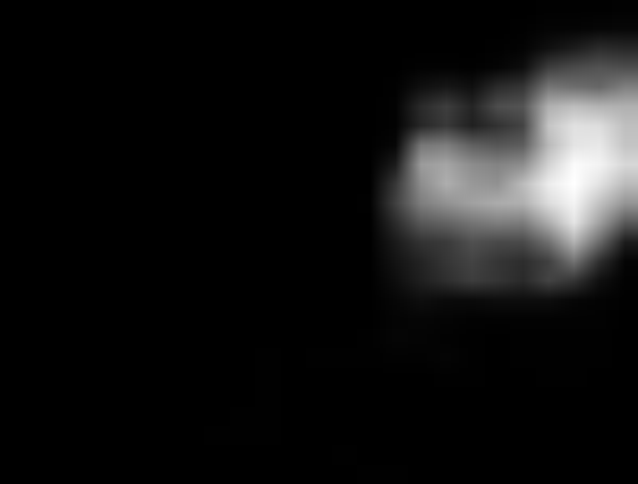}}
\subfigure[]{\includegraphics[width=.225\columnwidth]{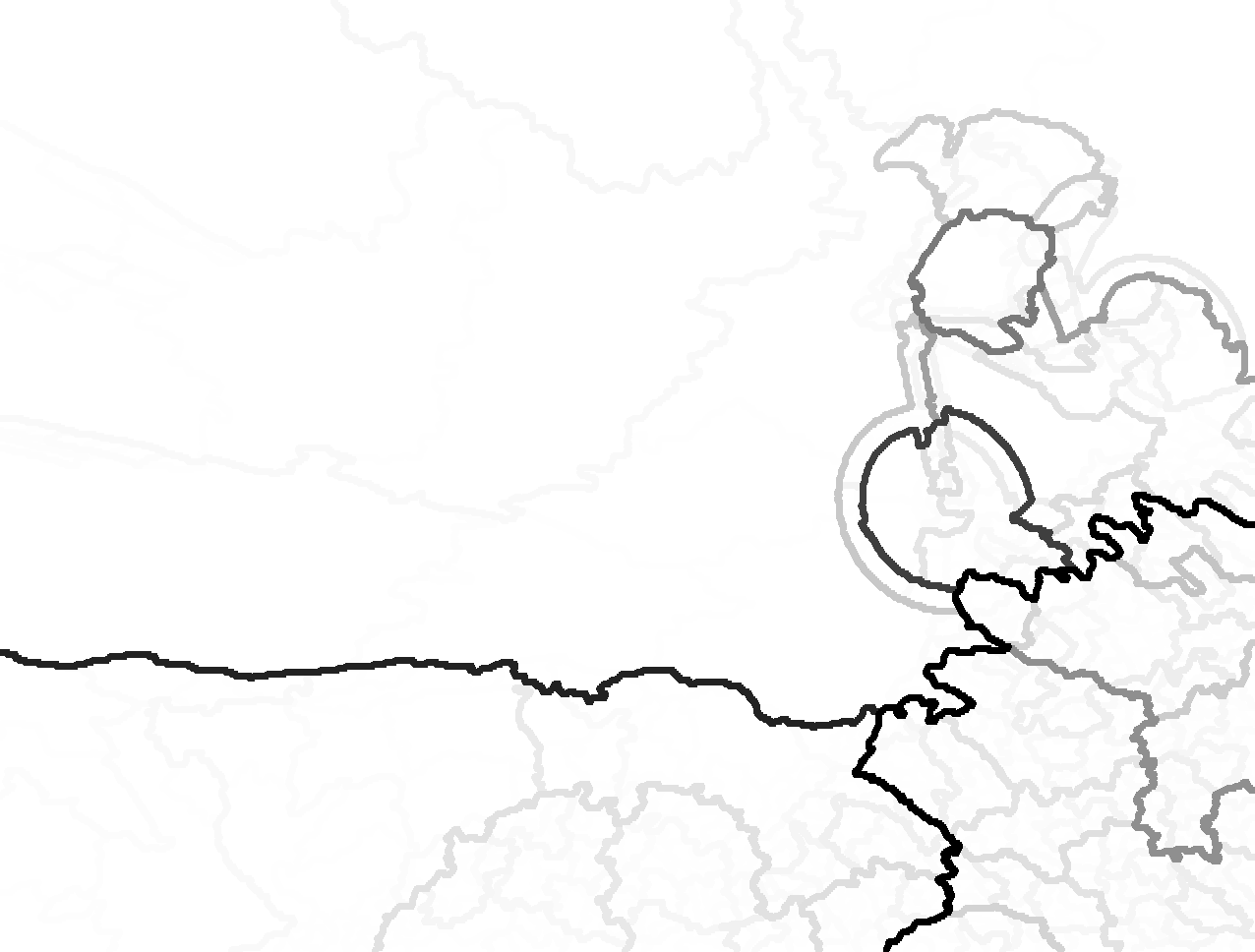}}
\subfigure[]{\includegraphics[width=.225\columnwidth]{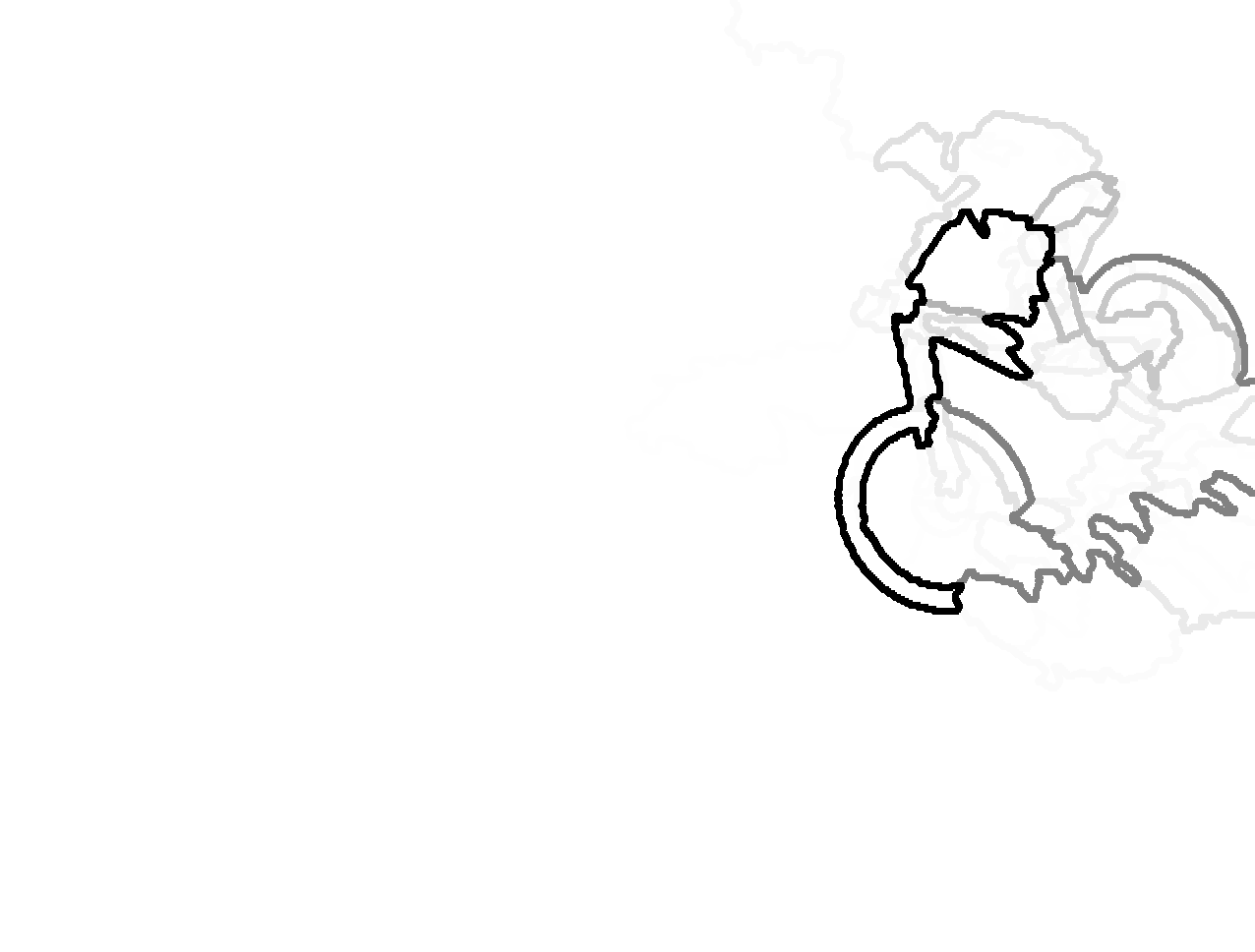}} \\
\squeezeup
\subfigure[]{\includegraphics[width=.325\columnwidth]{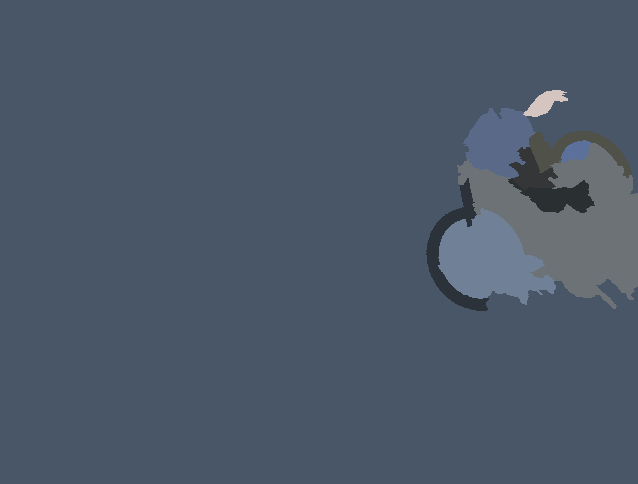}}
\subfigure[]{\includegraphics[width=.325\columnwidth]{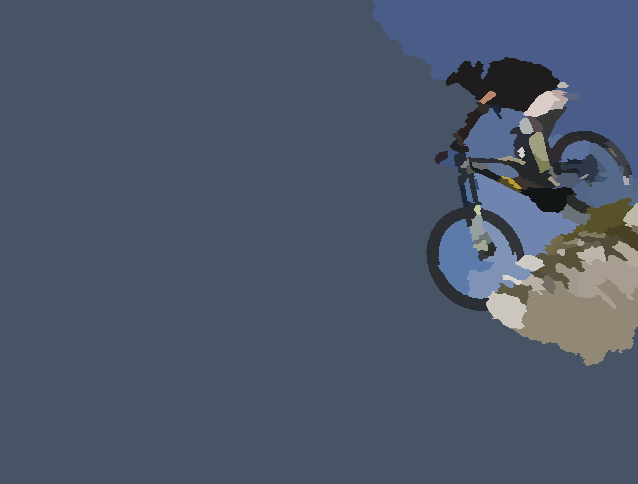}}
\subfigure[]{\includegraphics[width=.325\columnwidth]{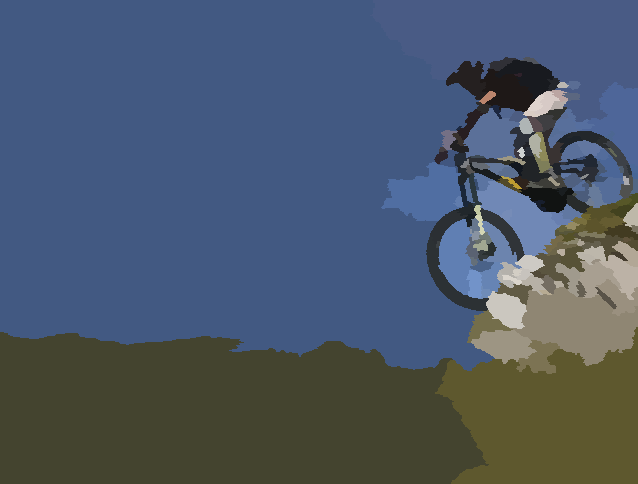}}
\squeezeup
\squeezeup
\caption{HFR focalisÈe sur la classe principale (``vÈlo") avec une $\probamap$ issue d'une mÈthode basÈe sur des CNN. (a) Image, (b) Heatmap de la classe principale, (c)(d)CCU de la LPES volumique et de la HFR avec information prior. (e)-(g) Segmentations obtenues avec la HFR - 10,100,200 rÈgions.}\label{Fig:CNN}
\end{center}
\squeezeup
\squeezeup
\end{figure}

Soit $E$ un objet d'une classe d'intÈrÍt, par exemple $E = {\text{``vÈlo"}}$, et $\I$ l'image ÈtudiÈe. On note $\theta_{E}$ la densitÈ de probabilitÈs associÈe ‡ $E$,  dÈfinie sur le domaine $D$ de $\I$ et obtenue par ailleurs. On note $\probamap(\I,\theta_{E})$ la carte de probabilitÈs associÈe, dans laquelle chaque pixel $p(x,y)$ de $\I$ prend comme valeur $\theta_{E}(x,y)$ sa probabilitÈ de faire partie de $E$. Etant donnÈe une telle information sur la position d'un objet d'intÈrÍt dans une image, on obtient une hiÈrarchie de segmentations focalisÈe sur cette rÈgion en modulant la distribution des marqueurs.

Si $\lambda$ est une densitÈ (uniforme ou non) dÈfinie sur $D$ pour distribuer des marqueurs, on choisit $\theta_{E} \lambda$ comme nouvelle densitÈ, favorisant ainsi l'Èmergence de contours parmi les rÈgions d'intÈrÍt. Si on considËre une rÈgion $\region$ de l'image, le nombre moyen de marqueurs tombant dans $\region$ est alors :

\begin{equation} \label{newnonuniform}
\Lambda_{E}(\region)=\int_{(x,y) \in \region} \theta_{E}(x,y) \lambda(x,y) \, \mathrm dx \mathrm dy
\end{equation}


De plus, cette approche peut aisÈment Ítre Ètendue au cas o˘ l'on veut tirer parti d'informations provenant de sources multiples. En effet, si 
$\theta_{E_1}$ et $\theta_{E_2}$ sont des densitÈs de probabilitÈs associÈes aux ÈvËnements $E_1$ et $E_2$, on peut combiner ces deux sources en utilisant comme nouvelle densitÈ $(\theta_{E_1}+\theta_{E_2})\lambda$.


\begin{figure}[H]
\begin{center}
\subfigure[]{\includegraphics[width=.275\columnwidth]{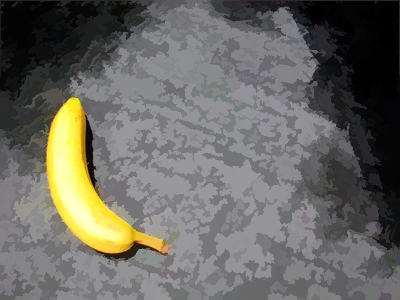}}
\subfigure[]{\includegraphics[width=.275\columnwidth]{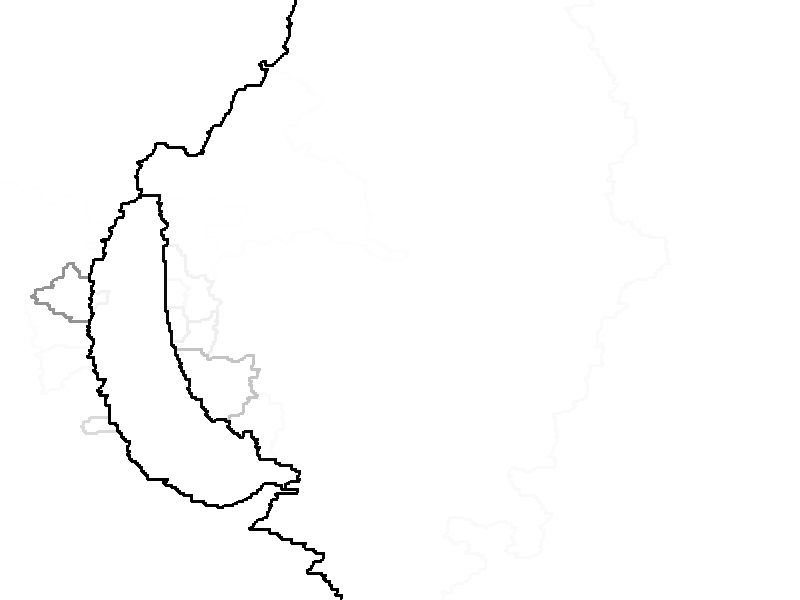}}
\subfigure[]{\includegraphics[width=.275\columnwidth]{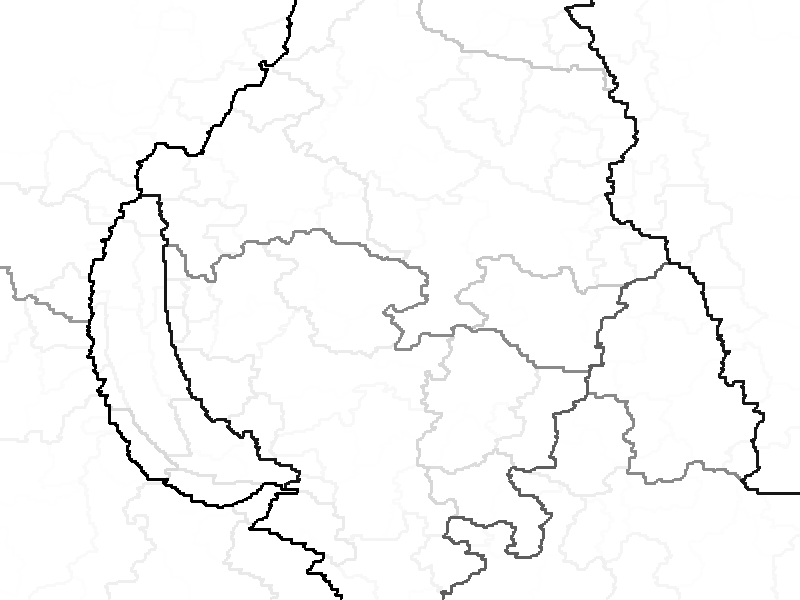}} \\
\squeezeup
\subfigure[]{\includegraphics[width=.275\columnwidth]{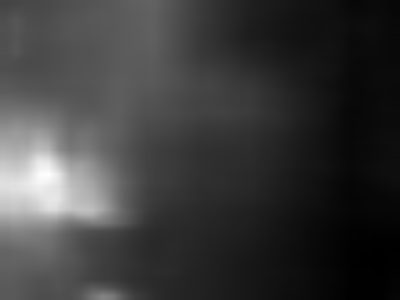}}
\subfigure[]{\includegraphics[width=.275\columnwidth]{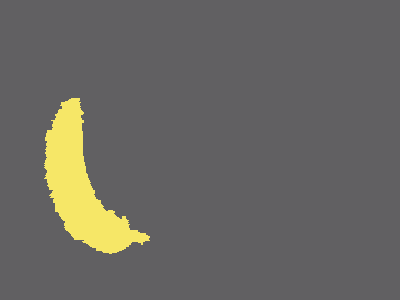}}
\subfigure[]{\includegraphics[width=.275\columnwidth]{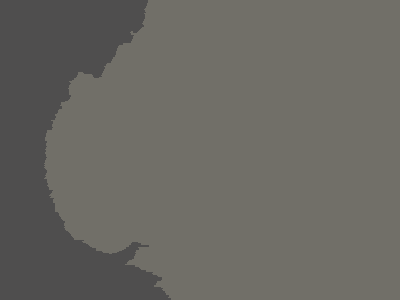}}
\squeezeup
\squeezeup
\caption{1\textsuperscript{er} exemple de HFR modulÈe (section \ref{ssec:varHFR}). (a) Image,  (b)(c) CCU de la HFR et de la LPES volumique, (d) Heatmap de la classe principale, (e)(f) Segmentations avec 2 rÈgions pour la HFR - ‡ gauche - et la LPES volumique - ‡ droite.}\label{Fig:CNN2}
\end{center}
\squeezeup
\squeezeup
\end{figure}

\subsection{Moduler la HFR en fonction des couples de rÈgions considÈrÈes}
\label{ssec:varHFR}
Si on veut favoriser l'Èmergence de certains contours, on peut moduler la densitÈ de marqueurs dans chaque rÈgion en prenant en compte la force des contours les sÈparant mais aussi leur position relative dans chacune de ces rÈgions.

Par exemple, pour tester la force du gradient sÈparant les deux rÈgions $\region_{s}$ et $\region_{t}$, on peut localement distribuer des marqueurs suivant une distribution $\chi(\region_{s},\region_{t}) \lambda$, avec $\chi(\region_{s},\region_{t})=\omega_{st}$. Cela correspond au cas classique de la LPES volumique, qui permet d'obtenir une hiÈrarchie prenant en compte ‡ la fois les surfaces des rÈgions et les contrastes entre elles.

On peut utiliser toute information spatiale a priori de faÁon similaire. En effet, en l'utilisant pour influencer la sortie de l'algorithme de segmentation, on peut vouloir choisir l'information pertinente ‡ mettre en valeur dans les segmentations rÈsultantes : le premier-plan, l'arriËre-plan ou les zones de transition entre les deux.

Par exemple, avoir plus de dÈtails dans les rÈgions de transition entre le premier-plan et l'arriËre-plan nous permet d'avoir plus de prÈcisions l‡ o˘ la limite entre les deux est justement floue. 
ConsidÈrons ce cas. On dÈfinit pour chaque couple de rÈgions $(\region_{s},\region_{t})$ un $\chi(\region_{s},\region_{t})$ pertinent. On veut alors que 
\newline
$\chi(\region_{s},\region_{t})$ soit petit si $\region_{s}$ et $\region_{t}$ sont au premier-plan ou ‡ l'arriËre-plan, et grand si $\region_{s}$ est ‡ l'arriËre-plan et $\region_{t}$ au premier plan (et vice-versa). On utilise :

\begin{equation}\label{eqNbMarkers2}
\left\{
\begin{array}{ll}
	\tilde{\lambda}= \chi \lambda \\
	\chi(\region_{s},\region_{t}) = \frac{\max(m(\region_{s}),m(\region_{t}))(1-\min(m(\region_{s}),m(\region_{t})))}{0.01+\sigma(\region_{s})\sigma(\region_{t})},
\end{array}
\right.
\end{equation}

$m(\region)$ (resp. $\sigma(\region)$) Ètant la moyenne normalisÈe (resp. l'Ècart-type normalisÈ) des valeurs des pixels dans la rÈgion $\region$ de $\probamap(\I)$.
Ainsi le nombre de marqueurs distribuÈs sera plus grand lorsque le contraste entre les rÈgions adjacentes sera grand (terme au numÈrateur) et lorsque ces rÈgions seront cohÈrentes (terme au dÈnominateur).

%
\squeezeup
\squeezeup
\section{Segmentation hiÈrarchique faiblement supervisÈe}
\label{sec:CNNprior}
\squeezeup

Il existe aujourd'hui de nombreux algorithmes permettant d'obtenir d'obtenir une localisation grossiËre de l'objet principal dans une image. On s'inspire de celle dÈcrite dans \cite{oquab15} pour ce faire. En utilisant le classifieur de type rÈseau de neurones convolutifs (CNN) VGG19 \cite{simonyan14} entraÓnÈ sur la base ImageNet \cite{deng09}, on commence par dÈterminer quelle est la classe principale dans l'image. Ce CNN prend comme entrÈe uniquement des images de taille $224\times224$ pixels. Une fois la classe de l'image dÈterminÈe, on peut alors, en changeant l'Èchelle de l'image d'un facteur $s \in \{0.5,0.7,1.0,1.4,2.0,2.8\}$, calculer pour des sous-fenÍtres de taille $224\times224$ de l'image, la probabilitÈ d'apparition de la classe principale. On obtient ainsi une carte de probabilitÈs de la classe principale pour chaque Èchelle. Par \textit{max-pooling} (``mise en commun en prenant le max"), on garde en mÈmoire le rÈsultat de l'Èchelle pour laquelle la probabilitÈ est la plus grande. La \textit{heatmap} (``carte de chaleur") ainsi gÈnÈrÈe prend en chaque pixel pour valeur la probabilitÈ que ce pixel appartienne ‡ la classe d'intÈrÍt. Elle peut alors Ítre utilisÈe en entrÈe de notre algorithme. De cette faÁon, nous avons ‡ notre disposition un outil pour se focaliser automatiquement sur la classe principale de chaque scËne. 

\begin{figure}[H]
\begin{center}
\subfigure[]{\includegraphics[width=.235\columnwidth]{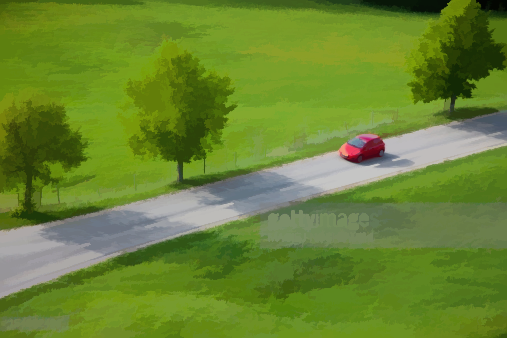}}
\subfigure[]{\includegraphics[width=.235\columnwidth]{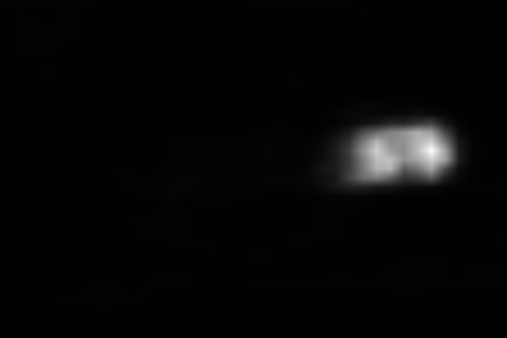}}
\subfigure[]{\includegraphics[width=.235\columnwidth]{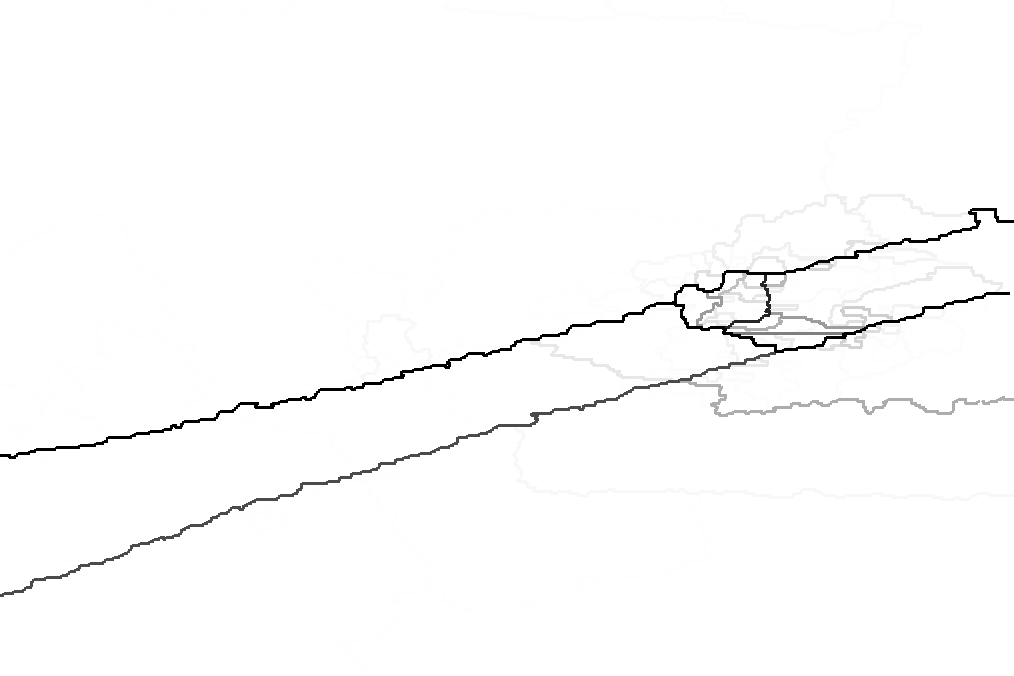}}
\subfigure[]{\includegraphics[width=.235\columnwidth]{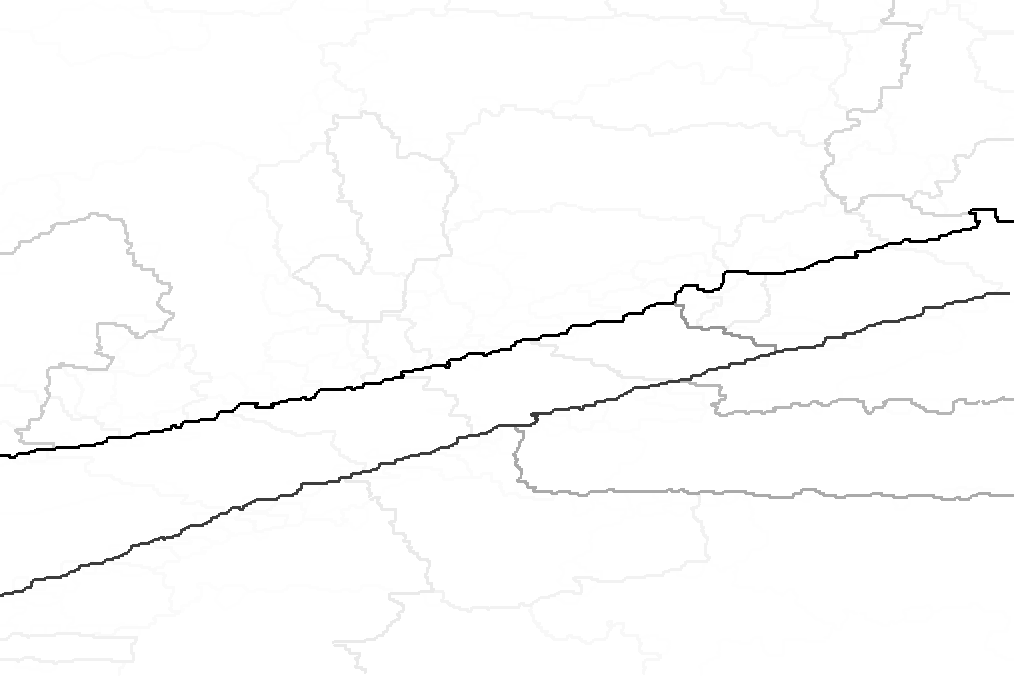}}
 \\
\squeezeup
\subfigure[]{\includegraphics[width=.235\columnwidth]{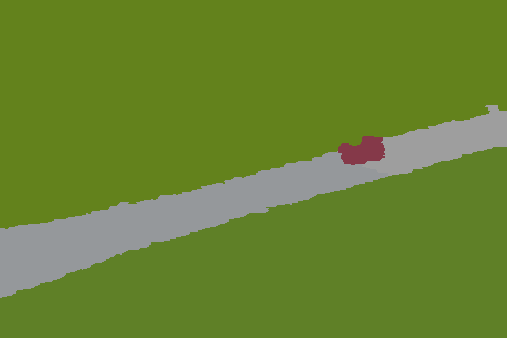}}
\subfigure[]{\includegraphics[width=.235\columnwidth]{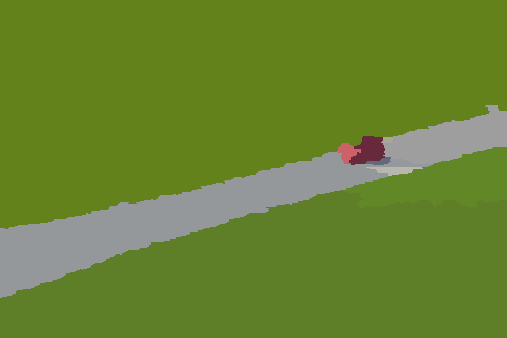}}
\subfigure[]{\includegraphics[width=.235\columnwidth]{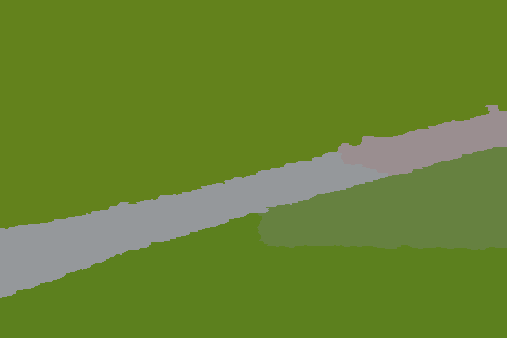}}
\subfigure[]{\includegraphics[width=.235\columnwidth]{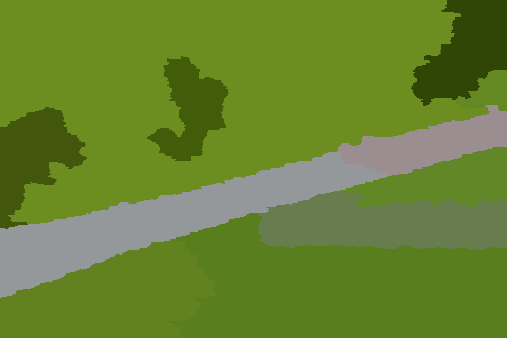}}
\squeezeup
\squeezeup
\caption{2\textsuperscript{Ëme} exemple de HFR modulÈe (section \ref{ssec:varHFR}). (a) Image, (b) Heatmap de la classe principale, (c)(d) CCU de la HFR et de la LPES volumique (e)-(h) Segmentations de 5 et 10 rÈgions pour la HFR - (e) et (f) - et la LPES volumique - (g) et (h).}\label{Fig:CNN3}
\end{center}
\end{figure}

Au final, la combinaison de cette mÈthode et de la HFR est trËs intÈressante puisqu'elle permet tout ‡ la fois de classifier l'image (i.e. d'en dÈterminer la classe principale), d'y localiser les ÈlÈments de la classe principale et d'obtenir une hiÈrarchie de segmentations focalisÈe sur ces ÈlÈments, i.e. une hiÈrarchie o˘ les contours appartenant ‡ ces ÈlÈments sont mis en exergue. Des rÈsultats visuels sont prÈsentÈs dans les figures \ref{Fig:CNN}, \ref{Fig:CNN2} et \ref{Fig:CNN3}.

%
%
\squeezeup
\squeezeup
\squeezeup
\squeezeup
\section{Conclusion}
\squeezeup
\label{sec:Conclusion}
Dans cet article, nous avons proposÈ un algorithme novateur et efficient de segmentation hiÈrarchique mettant l'accent sur des rÈgions d'intÈrÍt d'une image en utilisant une information exogËne obtenue sur celle-ci. La grande variÈtÈ de possibles sources d'information exogËne rend notre mÈthode extrÍmement versatile, et nous l'avons ici illustrÈe sur le problËme de segmentation hiÈrarchique faiblement supervisÈ. Pour aller plus loin, on pourrait chercher ‡ Ètendre ce travail ‡ des vidÈos.

\squeezeup
\squeezeup
\bibliographystyle{splncs03}
\bibliography{SegmentationBib}

%
%
%

\end{document}